%% file: paper.tex
\documentclass{article}

\usepackage{iclr2021_conference,times}
\input{math_commands.tex}

\iclrfinalcopy

\usepackage[utf8]{inputenc} 
\usepackage[T1]{fontenc}    
\usepackage{url}            
\usepackage{booktabs}       
\usepackage{amsfonts}       
\usepackage{nicefrac}       
\usepackage{microtype}      

\usepackage{amsmath}
\usepackage{amsthm}
\usepackage{graphicx}
\usepackage{caption}
\usepackage{subcaption}
\graphicspath{ {images/} }
\usepackage{mathtools}
\usepackage{xcolor}
\definecolor{dark-gray}{gray}{.35}
\definecolor{myorange}{RGB}{246, 164, 16}
\definecolor{mygreen}{RGB}{1, 100, 3}
\usepackage{footnote}
\makesavenoteenv{table}
\makesavenoteenv{tabular}
\usepackage{cleveref}[2012/02/15]
\crefformat{footnote}{#2\footnotemark[#1]#3}

\title{Gradient penalty from a maximum margin perspective}

\author{Alexia Jolicoeur-Martineau \\
  Department of Computer Science\\
  University of Montreal
  \And
  Ioannis Mitliagkas \\
  Department of Computer Science\\
  University of Montreal
}

\begin{document}

\maketitle

\begin{abstract}
A popular heuristic for improved performance in Generative adversarial networks (GANs) is to use some form of gradient penalty on the discriminator. This gradient penalty was originally motivated by a Wasserstein distance formulation. However, the use of gradient penalty in other GAN formulations is not well motivated. We present a unifying framework of expected margin maximization and show that a wide range of gradient-penalized GANs (e.g., Wasserstein, Standard, Least-Squares, and Hinge GANs) can be derived from this framework. Our results imply that employing gradient penalties induces a large-margin classifier (thus, a large-margin discriminator in GANs). We describe how expected margin maximization helps reduce vanishing gradients at fake (generated) samples, a known problem in GANs. From this framework, we derive a new $L^\infty$ gradient norm penalty with Hinge loss which generally produces equally good (or better) generated output in GANs than $L^2$-norm penalties (based on the Fréchet Inception Distance).
\end{abstract}

\section{Introduction}


    Generative adversarial networks (GANs) \citep{GAN} are a very successful class of generative models.
    Their most common formulation involves a game played between two competing neural networks, the discriminator $D$ and the generator $G$.
    $D$ is a classifier trained to distinguish real from fake examples, while $G$ is trained to generate fake examples that will confuse $D$ into recognizing them as real.
    When the discriminator's objective is maximized, it yields the value of a specific divergence (i.e., a distance between probability distributions) between the distributions of real and fake examples. The generator then aims to minimize that divergence (although this interpretation is not perfect; see \citet{jolicoeur2018beyonddivergence}).
    
    Importantly, many GANs apply some form of gradient norm penalty to the discriminator \citep{WGAN-GP,fedus2017many,mescheder2018training,karras2019style}. Gradient norm penalty has been widely adopted by the GAN community as a useful heuristic to improve the stability of GANs and the quality of the generated outputs. This penalty was originally motivated by a Wasserstein distance formulation in \citet{WGAN-GP}. However, its use in other GAN formulations is not well motivated. Given its success, one might wonder how one could derive an arbitrary GAN formulation with a gradient penalty?
    
    In this paper, we derive a framework which shows that gradient penalty arises in GANs from using a maximum margin classifier as discriminator. We then use this framework to better understand GANs and devise better gradient penalties. 
    
    The main contributions of this paper are:
	\begin{enumerate}
	    \item A unifying framework of expected margin maximization and showing that gradient-penalized versions of most discriminator/classifier loss functions (Wasserstein, Cross-entropy, Least-Squares, Hinge-Loss) can be derived from this framework.
	    \item A new method derived from our framework, a $L^\infty$ gradient norm penalty with Hinge function. We hypothesize and show that this method works well in GAN.
	    \item We describe how margin maximization (and thereby gradient penalties) helps reduce vanishing gradients at fake (generated) samples, a known problem in many GANs.
	    \item We derive the margins of Relativistic paired and average GANs \citep{jolicoeur2018relativistic,jolicoeur2019relativistic}.
	\end{enumerate}
	
	The paper is organized as follows. In Section~\ref{sec:2}, we show how gradient penalty arises from the Wasserstein distance in the GAN literature. In Section~\ref{sec:3}, we explain the concept behind maximum-margin classifiers (MMCs) and how they lead to some form of gradient penalty. In Section~\ref{sec:4}, we present our generalized framework of maximum-margin classification and experimentally validate it. In Section~\ref{sec:5}, we discuss of the implications of this framework on GANs and hypothesize that $L^1$-norm margins may lead to more robust classifiers. Finally, in Section~\ref{sec:6}, we provide experiments to test the different GANs resulting from our framework. Note that due to space constraints, we relegated the derivations of the margins of Relativistic GANs to Appendix~\ref{sec:5.5}.

\section{Gradient Penalty from the GAN Literature}
\label{sec:2}
\subsection{Notation}
\label{sec:notation}

We focus on binary classifiers. Let $f$ be the classifier and $(x,y) \sim \mathbb{D}$ the distribution (of a dataset $D$) with $n$ data samples $x$ and labels $y$. As per SVM literature, $y=1$ when $x$ is sampled from class 1 and $y=-1$ when $x$ is sampled from class 2. Furthermore, we denote $x_1 = x | (y = 1) \sim \mathbb{P}$ and $x_2 = x | (y = -1) \sim \mathbb{Q}$ as the data samples from class 1 and class 2 respectively (with distributions $\mathbb{P}$ and $\mathbb{Q}$). When discussing GANs, $x_1 \sim \mathbb{P}$ (class 1) refer to real data samples and $x_2 \sim \mathbb{Q}$ (class 2) refer to fake data samples (produced by the generator). The $L^{\infty}$-norm is defined as: $|| x ||_\infty = \max(|x_1|,|x_2|,\ldots,|x_k|)$.

The critic ($C$) is the discriminator ($D$) before applying any activation function (i.e., $D(x)=a(C(x))$, where $a$ is the activation function).
For consistency with existing literature, we will generally refer to the critic rather than the discriminator.

\subsection{GANs}

GANs can be formulated in the following way:
\begin{align}\label{eqn:3}
&\max_{C: \mathcal{X} \to \mathbb{R}} \mathbb{E}_{x_1 \sim \mathbb{P}}\left[ f_1(C(x_1)) \right] + \mathbb{E}_{z \sim \mathbb{Z}} \left[ f_2(C(G(z))) \right], \\
&\min_{G: Z \to \mathcal{X}} \mathbb{E}_{z \sim \mathbb{Z}} \left[ f_3(C(G(z))) \right],
\end{align}\label{eqn:4}
where $f_1, f_2, f_3:\mathbb{R} \to \mathbb{R}$, $\mathbb{P}$ is the distribution of real data with support $\mathcal{X}$,  $\mathbb{Z}$ is a multivariate normal distribution with support $Z=\mathbb{R}$, $C(x)$ is the critic evaluated at $x$, $G(z)$ is the generator evaluated at $z$, and $G(z) \sim \mathbb{Q}$, where $\mathbb{Q}$ is the distribution of fake data. 

Many variants exist; to name a few: Standard GAN (SGAN) \citep{GAN} corresponds to $f_1(z)=\log(\text{sigmoid}(z))$, $f_2(z)=\log(\text{sigmoid}(-z))$, and $f_3(z)=-f_1(z)$. Least-Squares GAN (LSGAN) \citep{LSGAN} corresponds to $f_1(z)=-(1-z)^2$, $f_2(z)=-(1+z)^2$, and $f_3(z)=-f_1(z)$. HingeGAN \citep{lim2017geometric} corresponds to $f_1(z)=-max(0,1-z)$, $f_2(z)=-max(0,1+z)$, and $f_3(z)=-z$.

\subsection{Integral Probability Metric Based GANs}

An important class of statistical divergences (distances between probability distributions) are Integral probability metrics (IPMs) \citep{muller1997integral}. IPMs are defined in the following way:
\[
IPM_{\mathcal{F}} (\mathbb{P} || \mathbb{Q}) = \sup_{C \in \mathcal{F}} \mathbb{E}_{x_1 \sim \mathbb{P}}[C(x_1)] - \mathbb{E}_{x_2 \sim \mathbb{Q}}[C(x_2)],
\]
where $\mathcal{F}$ is a class of real-valued functions.

A widely used IPM is the Wasserstein's distance ($W_1$), which focuses on the class of all 1-Lipschitz functions. This corresponds to the set of functions $C$ such that $\frac{C(x_1)-C(x_2)}{d(x_1,x_2)} \leq 1$ for all $x_1$,$x_2$, where $d(x_1,x_2)$ is a metric. $W_1$ also has a primal form which can be written in the following way:
\[ W_1 (\mathbb{P}, \mathbb{Q}):= \inf_{\pi \in \Pi (\mathbb{P}, \mathbb{Q})} \int_{M \times M} d(x_1, x_2) \, \mathrm{d} \pi (x_1, x_2),\]
where $\Pi (\mathbb{P}, \mathbb{Q})$ is the set of all distributions with marginals $\mathbb{P}$ and $\mathbb{Q}$ and we call $\pi$ a coupling.

The Wasserstein distance has been highly popular in GANs due to the fact that it provides good gradient for the generator in GANs which allows more stable training.

IPM-based GANs \citep{WGAN, WGAN-GP} attempt to solve the following problem \[ \min_{G} \max_{C \in \mathcal{F}} \mathbb{E}_{x_2 \sim \mathbb{P}}[C(x_1)] - \mathbb{E}_{z \sim \mathbb{Z}}[C(G(z))].\]

\subsection{Gradient Penalty as a way to estimate the Wasserstein Distance}
\label{sec:GANs}

To estimate the Wasserstein distance using its dual form (as a IPM), one need to enforce the 1-Lipschitz property on the critic. \citet{WGAN-GP} showed that one could impose a gradient penalty, rather than clamping the weights as originally done \citep{WGAN}, and that this led to better GANs. More specifically, they showed that if the optimal critic $f^{*}(x)$ is differentiable everywhere and that $\hat{x} = \alpha x_1 + (1-\alpha)x_2$ for $0 \leq \alpha \leq 1$, we have that $||\nabla C^{*}(\hat{x})||_2 = 1$ almost everywhere for all pair $(x_1,x_2)$ which comes from the optimal coupling $\pi^{*}$. 

Sampling from the optimal coupling is difficult so they suggested to softly penalize $\mathbb{E}_{\tilde{x}}{(||\nabla_{\tilde{x}} C(\tilde{x})||_2-1)^2}$, where $\tilde{x} =  \alpha x_1 + (1-\alpha)x_2$, $\alpha \sim U(0,1)$, $x_1 \sim \mathbb{P}$, and $x_2 \sim \mathbb{Q}$. They called this approach Wasserstein GAN with gradient-penalty (WGAN-GP). However, note that this approach does not necessarily estimate the Wasserstein distance since we are not sampling from $\pi^{*}$ and $f^{*}$ does not need to be differentiable everywhere \citep{petzka2017regularization}.

Of importance, gradient norm penalties of the form $\mathbb{E}_{x}{(||\nabla_{x} D(x)||_2-\delta)^2}$, for some $\delta \in \mathbb{R}$ are very popular in GANs. Remember that $D(x)=a(C(x))$; in the case of IPM-based-GANs, we have that $D(x)=C(x)$. It has been shown that the GP-1 penalty ($\delta=1$), as in WGAN-GP, also improves the performance of non-IPM-based GANs \citep{ManyPaths}. Another successful variant is GP-0 ($\delta=0$ and $x \sim \mathbb{P}$) \citep{mescheder2018training,karras2019style}. Although there are explanations to why gradient penalties may be helpful \citep{mescheder2018training, kodali2017convergence, WGAN-GP}, the theory is still lacking.

\section{Maximum-Margin Classifiers}
\label{sec:3}

In this section, we define the concepts behind maximum-margin classifiers (MMCs) and show how it leads to a gradient penalty.

\subsection{Decision Boundary and Margin}

The \em decision boundary \em of a classifier is defined as the set of points $x_0$ such that $f(x_0)=0$.

The margin is either defined as
i) the minimum distance between a sample and the boundary, or
ii) the minimum distance between the \em closest sample \em to the boundary and the boundary.
The former thus corresponds to the \em margin of a sample \em and the latter corresponds to the \em margin of a dataset \em.
In order to disambiguate the two cases, we refer to the former as the \em margin \em and the latter as the \em minimum margin\em.

\subsection{Geometric Margin and Gradient Penalty}

The first step towards obtaining a  MMC is to define the $L^p$-norm margin:
\begin{align}\label{eq:10}
\gamma(x) = &\min_{x_0} || x_0 - x ||_p \hspace{5pt} \text{ s.t. } \hspace{5pt} f(x_0)=0
\end{align}

With a linear classifier (i.e., $f(x) = w^T x$), there is a close form solution. However, the formulation of the $L^p$-norm margin \eqref{eq:10} has no closed form for arbitrary non-linear classifiers. A way to derive an approximation of the margin is to use Taylor's approximation before solving the problem (as done by \citet{matyasko2017margin} and \citet{elsayed2018large}):
\begin{align*}
\gamma_p(x) &= \min_{r} || r ||_p \hspace{5pt} \text{ s.t. } \hspace{5pt} f(x+r)=0 \\
&\approx \min_{r} || r ||_p \hspace{5pt} \text{ s.t. } \hspace{5pt} f(x)+\nabla_x f(x)^T r=0 \\
&=\frac{|f(x)|}{|| \nabla_x f(x) ||_q},
\end{align*}
where $||\cdot||_q$ is the dual norm \citep{boyd2004convex} of $||\cdot||_p$. By Hölder's inequality \citep{holder1889ueber, rogers1888extension}, we have that $1/p + 1/q=1$. This means that if $p=2$, we still get $q=2$; if $p=\infty$, we get $q=1$; if $p=1$, we get $q=\infty$.

The goal of MMCs is to maximize a margin, but also to obtain a classifier. To do so, we simply replace $\alpha(x)=|f(x)|$ by $\widetilde{\alpha}(x,y) = y f(x)$. We call $\widetilde{\alpha}$ the \em functional margin\em. After replacement, we obtain the \em geometric margin\em:
\[\widetilde{\gamma}(x,y) = \frac{y f(x)}{|| \nabla_x f(x) ||_q}\]

If $p=2$ and $f(x)$ is linear, this leads to the same geometric margin used in Support Vector-Machines (SVMs). \citet{matyasko2017margin} used this result to generalize Soft-SVMs to arbitrary classifiers by simply penalizing the $L^p$-norm of the gradient rather than penalizing the $L^p$-norm of the model's weights (as done in SVMs). Meanwhile, \citet{elsayed2018large} used this result in a multi-class setting and maximized the geometric margin directly.

\section{Generalized framework of Maximum-margin classification}
\label{sec:4}

\subsection{Framework}

Here we show how to generalize the idea behind maximizing the geometric margin into arbitrary loss functions with a gradient penalty. Directly maximizing the geometric margin is an ill-posed problem. The numerator and denominator are dependent on one another; increasing the functional margin also increases the norm of the gradient (and vice-versa). Thereby, there are infinite solutions which maximize the geometric margin. For this reason, the common approach (as in SVM literature; see \citet{cortes1995support}) is to: i) constrain the numerator and minimize the denominator, or ii) constrain the denominator and maximize the numerator.

Approach i) consists of minimizing the denominator and constraining the numerator using the following formulation:
\begin{align}\label{eqn:2} 
\min_{f} { ||\nabla_x f(x)||_p} \hspace{5pt} \text{ s.t. } \hspace{5pt} yf(x) \ge 1 \hspace{2pt}\forall\hspace{2pt} (x,y) \in D
\end{align}

The main limitation of this approach is that it only works when the data are separable. However, if we take the opposite approach of maximizing a function of $y f(x)$ and constraining the denominator $|| f(x) ||_2$, we can still solve the problem with non-separable data. This corresponds to approach ii):
\begin{align}
 \max_{f} \mathbb{E}_{(x,y)\sim \mathbb{D}} \left[y f(x)\right] \hspace{1pt} \text{ s.t. } \hspace{1pt} ||\nabla_x f(x)||_q \leq 1 \text{ or } ||\nabla_x f(x)||_q = 1.
\end{align}

The constraint chosen can be enforced by either i) using a KKT multiplier \citep{kuhn1951nonlinear, karush1939minima} or ii) approximately imposing it with a soft-penalty. Furthermore, one can use any margin-based loss function rather than directly maximize $yf(x)$. Thus, we can generalize this idea by using the following formulation:
\begin{align}\label{eqn:6}
\min_f \mathbb{E}_{(x,y)\sim \mathbb{D}}\left[L(yf(x)) + \lambda g(||\nabla_x f(x)||_q)\right].
\end{align}
where $L,g:\mathbb{R}\to \mathbb{R}$ and $\lambda$ is a scalar penalty term. There are many potential choices of $L$ and $g$ which we can use. 

If $L$ is chosen to be the hinge function (i.e., $L(z)=max(0,1-z)$),
we ignore samples far from the boundary (as in Hard-Margin SVMs). For general choices of $L$, every sample may influence the solution. The identity function $L(z)=z$, cross entropy with sigmoid activation $L(z)=-\log(\text{sigmoid}(z)))$ and least-squares $L(z)=(1-z)^2$ are also valid choices.

A standard choice of $g$ is $g(z)=(z^2-1)$. This corresponds to constraining $|| \nabla_x f(x) ||_2^2 = 1$ or $|| \nabla_x f(x) ||_2^2 \leq 1$ for all $x$ (by KKT conditions). As an alternative, we can also consider soft constraints of the form $g(z)=(z-1)^2$ or $g(z)=max(0,z-1)$. The first function enforces a soft equality constraint so that $z\approx 1$  while the second function enforces a soft inequality constraint so that $z \leq 1$. Soft constraints are useful if the goal is not to obtain the maximum margin solution but to obtain a solution that leads to a large-enough margin.

Of importance, MMCs can be seen as a generalization of Support Vector Machines (SVMs). When $p=2$ and $f$ is linear ($f(x)=w^T x$), equation \ref{eqn:2} corresponds exactly to Hard-Margin SVMs and equation \ref{eqn:6} with $L(z)=max(0,1-z)$ and $g(z)=(z^2-1)$ corresponds exactly to Soft-Margin SVMs \citep{cortes1995support}.

\subsection{Experimental evidence of large margin from gradient penalties}

We ran experiments to empirically show that gradient-penalized classifiers (trained to optimize equation \ref{eqn:6}) maximize the expected margin. We used the swiss-roll dataset \citep{marsland2015machine} to obtain two classes (one is the swiss-roll and one is the swiss-roll scaled by 1.5).  The results are shown in Table~\ref{tab:9} (Details of the experiments are in Appendix~\ref{sec:111}).

\begin{table}[!ht]
	\caption{Expected $L^p$ Margin for different types of gradient penalties (or none). Classifier was trained on the swiss-roll dataset with a cross-entropy loss function.}
	\label{tab:9}
	\centering
	\begin{tabular}{cccc}
		\toprule
		& \multicolumn{3}{c}{Expected $L^p$ Margin} \\
		Type of gradient penalty & $p=2$ & $p=1$ & $p=\infty$ \\
		\cmidrule(){1-4}
		No gradient penalty & .27 & .25 & .24 \\
		\cmidrule(){1-4}
		$g(z)=(z-1)^2$ \\
		\cmidrule(){1-4}
        $L^2$ gradient penalty ($L^2$ margin) & .62 & .75 & .64 \\
        $L^\infty$ gradient penalty ($L^1$ margin) & .43 & .58 & .33 \\
        $L^1$ gradient penalty ($L^\infty$ margin) & .69 & .85 & .60 \\
        \cmidrule(){1-4}
        $g(z)=\text{max}(0,z-1)$ \\
        \cmidrule(){1-4}
        $L^2$ gradient penalty ($L^2$ margin) & .43 & .53 & .37 \\
        $L^\infty$ gradient penalty ($L^1$ margin) & .41 & .56 & .31 \\
        $L^1$ gradient penalty ($L^\infty$ margin) & .43 & .44 & .42 \\
		\bottomrule
	\end{tabular}
\end{table}

We observe that we obtain much larger expected margins (generally 2 to 3 times bigger) when using a gradient penalty; this is true for all types of gradient penalties.

\section{Implications of the Maximum Margin Framework on GANs}
\label{sec:5}

\subsection{GANs can be derived from the MMC Framework}
\label{sec:5.1}

Although not immediately clear given the different notations, let $f(x)=C(x)$ and we have:
\begin{align*}
\mathbb{E}_{(x,y)\sim \mathbb{D}}\left[L(yf(x))\right] = \mathbb{E}_{x_1 \sim \mathbb{P}}[L(C(x_1))] + \mathbb{E}_{z \sim \mathbb{Z}}[L(-C(G(z)))].
\end{align*}
Thus, the objective functions of the discriminator/critic in many penalized GANs are equivalent to the ones from MMCs based on \eqref{eqn:6}. We also have that $L(z)=\log(\text{sigmoid}(z))$ corresponds to SGAN, $L(z)=(1-z)^2$ corresponds to LSGAN, and $L(z)=max(0,1-z)$ corresponds to HingeGAN. When $g(z)=(z-1)^2$, we also have that $L(z)=z$ corresponds to WGAN-GP. Thus, most $L^p$-norm gradient penalized GANs imply that the discriminator approximately maximize an expected $L^q$-norm margin.

\subsection{Why do Maximum-Margin Classifiers make good GAN Discriminators/Critics?}
\label{sec:5.3}

To show that maximizing an expected margin leads to better GANs, we prove the following statements:
\begin{enumerate}
    \item classifier maximizes an expected margin $\iff$ classifier has a fixed Lipschitz constant
    \item MMC with a fixed Lipschitz constant $\implies$ better gradients at fake samples
   \item better gradients at fake samples $\implies$ stable GAN training.
\end{enumerate}

\subsubsection{Equivalence between gradient norm constraints and Lipschitz functions}
\label{sec:5.2}

As stated in Section~\ref{sec:GANs}, the WGAN-GP approach of softly enforcing $|| \nabla_{\tilde{x}} f(\tilde{x}) ||_2 \approx 1$ at all interpolations between real and fake samples does not ensure that we estimate the Wasserstein distance ($W_1$).
On the other hand, we show here that enforcing $|| \nabla_x f(x) ||_2 \leq 1$ is sufficient in order to estimate $W_1$.

Assuming $d(x_1,x_2)$ is a $L^p$-norm, $p \ge 2$ and $f(x)$ is differentiable, we have that:
\begin{align*}
|| \nabla f(x) ||_p \leq K \iff f \text{ is K-Lipschitz on $L^p$}.
\end{align*} See appendix for the proof. \citet{adler2018banach} showed a similar result on dual norms. 

This suggests that, in order to work on the set of Lipschitz functions, we should enforce that $|| \nabla_x f(x) || \leq 1$ for all $x$.
This can be done, through \eqref{eqn:6}, by choosing $g(z)=(z^2-1)$ or, in approximation (using a soft-constraint), by choosing $g(z)=\max(0,z-1)$. \citet{petzka2017regularization} suggested using a similar function (the square hinge) in order to only penalize gradient norms above 1.

If we let $L(z)=z$ and $g(z)=\max(0,z-1)$, we have an IPM over all Lipschitz functions. thus, we effectively approximate $W_1$. This means that $W_1$ can be found through maximizing a geometric margin. Meanwhile, WGAN-GP only leads to a lower bound on $W_1$.

Importantly, most successful GANs \citep{brock2018large,karras2019style,karras2017progressive} either enforce the 1-Lipschitz property using Spectral normalization \citep{miyato2018spectral} or use some form of gradient norm penalty \citep{WGAN-GP,mescheder2018training}. Since 1-Lipschitz is equivalent to enforcing a gradient norm constraint (as shown above), we have that most successful GANs effectively train a discriminator/critic to maximize a geometric margin.

The above shows that training an MMC based on equation~(\ref{eqn:6}) is equivalent to training a classifier with a fixed Lipschitz constant.

\subsubsection{MMC leads to better gradient at fake samples}

\begin{figure}[!ht]
	\centering
	\begin{subfigure}[t]{0.5\textwidth}
    	\centering\includegraphics[scale=0.45]{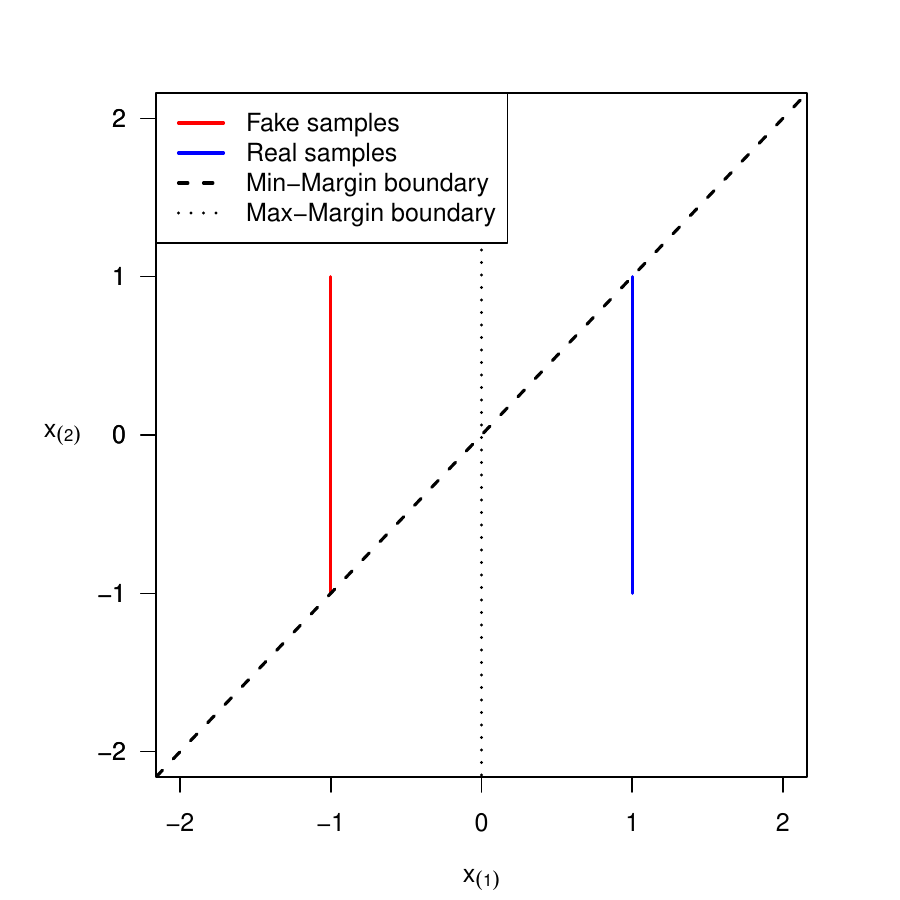}
    	\caption{}
    	\label{fig:fig1}
	\end{subfigure}%
	\begin{subfigure}[t]{0.5\textwidth}
    	\centering\includegraphics[scale=0.45]{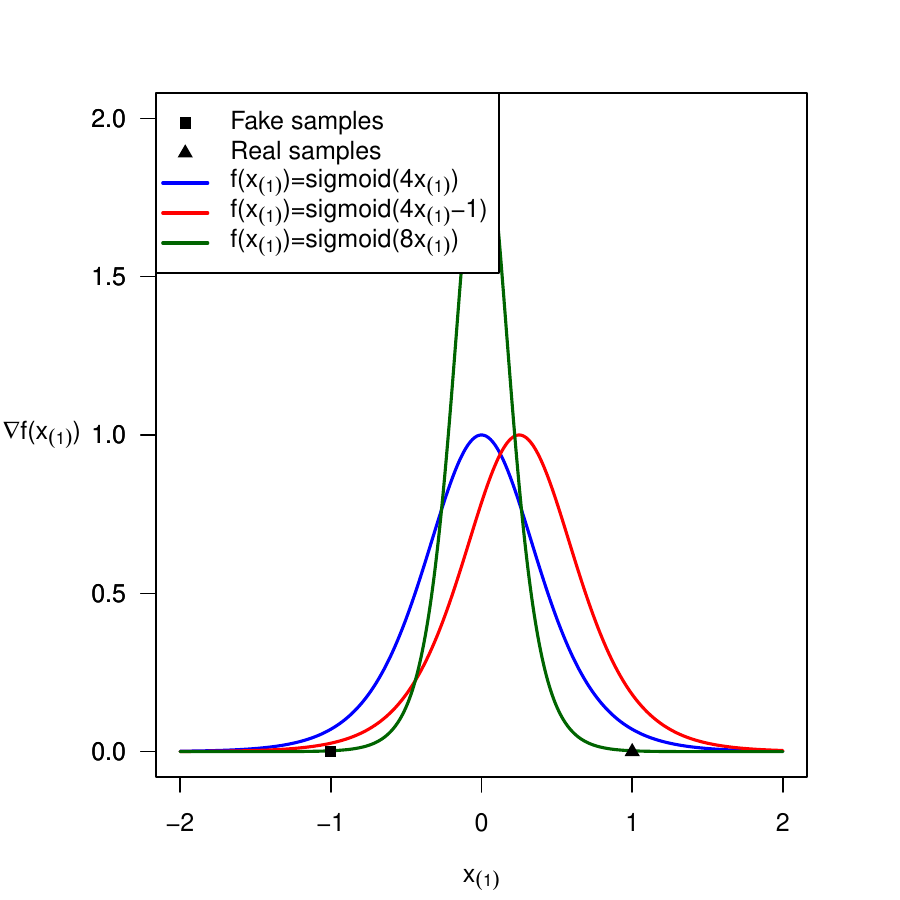}
    	\caption{}
    	\label{fig:fig2}
	\end{subfigure}
	\caption{a) Two-dimensional GAN example with different choices of boundaries, b) $\nabla f(x_{(1)})$ at different values of $x_{(1)}$ for the two-dimensional example assuming a sigmoid function.}
\end{figure}

Consider a simple two-dimensional example where $x=(x_{(1)},x_{(2)})$. Let real data (class 1) be uniformly distributed on the line between $(1,-1)$ and $(1,1)$. Let fake data (class 2) be uniformly distributed on the line between $(-1,-1)$ and $(-1,1)$. This is represented by Figure~\ref{fig:fig1}.  Clearly, the maximum-margin boundary is the line $x_{(1)}=0$ and any classifier should learn to ignore $x_{(2)}$. 

Consider a non-linear classifier of the form $f(x)=\text{sigmoid}(w_1 x_{(1)}+w_0)$ (See Figure~\ref{fig:fig2}). To ensure we obtain an MMC, we need to enforce $||\nabla_{x} f(x)|| \leq K$.

The best classifier with Lipschitz constant $K=1$ is obtained by choosing $w_1 = 4$. The maximum-margin boundary is at $x_{(1)}=0$ (which we get by taking $w_0=0$; blue curve in Figure~\ref{fig:fig2}); for this choice, we have that $f(x_{r})=.02$ and $f(x_{f})=.98$ for real ($x_r$) and fake ($x_f$) samples respectively. Meanwhile, if we take a slightly worse margin with boundary at $x_{(1)} = \frac{1}{4}$ (equivalent to choosing $w_0=-1$; red curve in Figure~\ref{fig:fig2}), we have that $f(x_{r})=.01$ and $f(x_{f})=.95$ for real and fake samples respectively. Thus, both solutions almost perfectly classify the samples. However, the optimal margin has gradient $.07$, while the worse margin has gradient $.03$ at fake samples; this is why maximizing a margin lead to similar signal for real and fake samples. Furthermore, if we had enforced a bigger Lipschitz constant ($K=2$), the best classifier would have been obtained with $w_1 = 8$ (green curve in Figure~\ref{fig:fig2}); this would have caused vanishing gradients at fake samples unless we had scaled up the learning up. Thus, for a fixed or decreasing learning rate, it is important to fix $K$ (and ideally to a small value) in order for the gradient signal to be strong at fake samples.

In summary, the maximum-margin discriminator provides a stronger signal at fake samples by preventing a sharp change in the discriminator (i.e., small gradient near real/fake data and large gradient between real and fake data) and centering the classifier so that the gradients at real and fake samples are similar. This further suggests that imposing a gradient penalty in the interpolation between real and fake data (as done in WGAN-GP) is most sensible to ensure that the gradient norm remains small between real and fake data. 

\subsubsection{Better gradients at fake samples implies stable GAN training}

In GANs, the dynamics of the game depends in great part on $\nabla_{x_{f}} f(x_{f})$ where $x_{f}$'s are samples from the fake, or generated, distribution.
This is because the generator only learns through the discriminator/critic and it uses $\nabla_{x_{f}} f(x_{f})$ in order to improve its objective function. Thus, for stable training with a fixed or decreasing learning rate, $||\nabla_{x_{f}} f(x_{f})||$ should not be too small.

The above means that, in order to get stable GAN training, we need to ensure that we obtain a solution with a stable non-zero gradient around fake samples. Thus, it is preferable to solve the penalized formulation from equation \eqref{eqn:6} and choose a large penalty term $\lambda$ in order to obtain a small-gradient solution. 

\subsection{Are certain Margins better than others?}
\label{sec:5.4}

It is well known that $L^p$-norms (with $p\ge 1$) are more sensitive to outliers as $p$ increases which is why many robust methods minimize the $L^1$-norm \citep{bloomfield1983least}. Furthermore, minimizing the $L^1$-norm loss results in a median estimator \citep{bloomfield1983least}. This suggests that  penalizing the $L^2$ gradient norm penalty ($p=2$) may not lead to the most robust classifier. We hypothesize that $L^\infty$ gradient norm penalties may improve robustness in comparison to $L^2$ gradient norm penalties since they correspond to maximizing $L^1$-norm margin.
In Section~\ref{sec:6}, we provide experimental evidence in support of our hypothesis.

\section{Experiments}
\label{sec:6}

Following our analysis and discussion in the previous sections, we hypothesized that $L^1$ margins, corresponding to a $L^\infty$ gradient norm penalty, would perform better than $L^2$ margins ($L^2$ gradient norm penalty).
As far as we know, researchers have not yet tried using a $L^\infty$ gradient norm penalty in GANs.
In addition, we showed that it would be more sensible to penalize violations of $||\nabla f(x)||_q \leq 1$ rather than $||\nabla f(x)||_q \approx 1$.

To test these hypotheses, we ran experiments on CIFAR-10 (a dataset of 60k images from 10 categories) \citep{krizhevsky2009learning} using HingeGAN ($L(z)=\max(0,1-z)$) and WGAN ($L(z)=z$) loss functions with $L^1$, $L^2$, $L^\infty$ gradient norm penalties. We enforce either $||\nabla f(x)||_q \approx 1$ using Least Squares (LS) $(g(z)=(z-1)^2)$ or $||\nabla f(x)||_q \leq 1$ using Hinge $(g(z)=\max(0,z-1))$. We used the standard hyperparameters: a learning rate (lr) of .0002, a batch size of 32, and the ADAM optimizer \citep{Adam} with parameters $(\alpha_1, \alpha_2)= (.50, .999)$ We used a DCGAN architecture \citep{DCGAN}. As per convention, we report the Fréchet Inception Distance (FID) \citep{heusel2017gans}; lower values correspond to better generated outputs (higher quality and diversity). As per convention, all 50k images from the training part of the dataset were used for training and to calculate the FID. We ran all experiments using seed 1 and with gradient penalty $\lambda=20$. Details on the architectures are in the Appendix. All models were trained using a single GPU. Code is available on \em https://github.com/AlexiaJM/MaximumMarginGANs\em. The results are shown in Table~\ref{tab:1}.

\begin{table}[!ht]
	\caption{Fréchet Inception Distance (FID) after 100k generator iterations on CIFAR-10.}
	\label{tab:1}
	\centering
	\begin{tabular}{ccc}
		\toprule
		$g(||\nabla_x f(x))||_q)$ & WGAN & HingeGAN \\
		\cmidrule(){1-3}
		$(||\nabla_x f(x))||_1-1)^2$ & 99.7 & 88.9 \\
		$\max(0,||\nabla_x f(x))||_1-1)$ & 65.6 &  77.3 \\
		\cmidrule(){1-3}
		$(||\nabla_x f(x))||_2-1)^2$ & 37.6 & 32.8 \\
		$\max(0,||\nabla_x f(x))||_2-1)$ & 37.8 & 33.9 \\
		\cmidrule(){1-3}
		$(||\nabla_x f(x))||_{\infty}-1)^2$ & 33.4 & 33.6 \\
		$\max(0,||\nabla_x f(x))||_{\infty}-1)$ & 36 &  \fontseries{b}\selectfont 27.1 \\
		\bottomrule
	\end{tabular}
\end{table}

Due to space constraint, we only show the previously stated experiments in Table~\ref{tab:1}. However, we also ran additional experiments on CIFAR-10 with 1) Relativistic paired and average HingeGAN, 2) $\beta=(0,.90)$, 3) the standard CNN architecture from \citet{miyato2018spectral}. Furthermore, we ran experiments on CAT \citep{cat} with 1) Standard CNN (in 32x32), and 2) DCGAN (in 64x64). These experiments correspond to Table~\ref{tab:2},~\ref{tab:3},~\ref{tab:4},~\ref{tab:5}, and~\ref{tab:6} from the appendix.

In all sets of experiments, we generally observed that we obtain smaller FIDs by using: i) a larger $q$ (as theorized), ii) the Hinge penalty to enforce an inequality gradient norm constraint (in both WGAN and HingeGAN), and iii) HingeGAN instead of WGAN. 

\section{Conclusion}
\label{sec:7}


This work provides a framework in which to derive MMCs that results in very effective GAN loss functions. In the future, this could be used to derive new gradient norm penalties which further improve the performance of GANs. Rather than trying to devise better ways of enforcing 1-Lipschitz, researchers may instead want to focus on constructing better MMCs (possibly by devising better margins).

This research shows a strong link between GANs with gradient penalties, Wasserstein's distance, and SVMs. Maximizing the minimum $L^2$-norm geometric margin, as done in SVMs, has been shown to lower bounds on the VC dimension which implies lower generalization error \citep{vapnik1998statistical,mount2015sure}. This paper may help researchers bridge the gap needed to derive PAC bounds on Wasserstein's distance and GANs/IPMs with gradient penalty. Furthermore, it may be of interest to theoreticians whether certain margins lead to lower bounds on the VC dimension.


\section{Acknowledgements}
This work was supported by Borealis AI through the Borealis AI Global Fellowship Award. We would also like to thank Compute Canada and Calcul Québec for the GPUs which were used in this work.
This work was also partially supported by the FRQNT new researcher program (2019-NC-257943), the NSERC Discovery grant (RGPIN-2019-06512), a startup grant by IVADO, a grant by Microsoft Research and a Canada CIFAR AI chair.

\bibliographystyle{unsrtnat}
\bibliography{paper}

\clearpage

\appendix
\section*{Appendices}
\addcontentsline{toc}{section}{Appendices}
\renewcommand{\thesubsection}{\Alph{subsection}}

\subsection{Details on experiments for Table 1}
\label{sec:111}

The classifier was a 4 layers fully-connected neural network with ReLU activation functions. It was trained for $10K$ iterations with the cross-entropy loss, the ADAM optimizer, a batch size of $256$, $\lambda=10$, and a learning rate of $.00005$. The margins (as in equation \ref{eq:10}) were estimated using Gradient Descent with the Augmented Lagrangian method \citep{hestenes1969multiplier} until $|f(x_0)|<.01$. We estimated the expected margin using the average margin from 256 random samples of both classes.

\subsection{Additional experiments}

Note that the \em smooth maximum \em is defined as
\[\text{smax}(x_{(1)},\ldots,x_{(k)})=\frac{\sum_{i=1}^{k} x_{(i)} e^{x_i}}{\sum_{i=1}^{k} e^{x_i}}.\]
We sometime use the smooth maximum as a smooth alternative to the $L^{\infty}$-norm margin; results are worse with it.

\begin{table}[!ht]
	\caption{FID after 100k generator iterations on CIFAR-10 using the same setting as Table1, but we are using Relativistic paired and average GANs.}
	\label{tab:2}
	\centering
	\begin{tabular}{ccc}
		\toprule
		$g(||\nabla_x f(x))||_q)$ & RpHinge & RaHinge \\
		\cmidrule(){1-3}
		$(||\nabla_x f(x))||_1-1)^2$ & 64.4 & 65.0 \\
		$\max(0,||\nabla_x f(x))||_1-1)$ & 60.3 &  68.5 \\
		\cmidrule(){1-3}
		$(||\nabla_x f(x))||_2-1)^2$ & 32.8 & 31.9 \\
		$\max(0,||\nabla_x f(x))||_2-1)$ & 32.6 & 35.0 \\
		\cmidrule(){1-3}
		$(||\nabla_x f(x))||_{\infty}-1)^2$ & 32.5 & 33.5 \\
		$\max(0,||\nabla_x f(x))||_{\infty}-1)$ & \fontseries{b}\selectfont 28.2 & 28.4 \\
		\cmidrule(){1-3}
		$(\text{smax}|\nabla_x f(x)|-1)^2$ & 133.7 & 124.0 \\
		$\max(0,\text{smax}|\nabla_x f(x)|-1)$ & 30.5 & 30.3 \\
		\bottomrule
	\end{tabular}
\end{table}

\begin{table}[!ht]
	\caption{FID after 100k generator iterations on CIFAR-10 using the same setting as Table1, but we are using Adam $\beta=(0, .90)$.}
	\label{tab:3}
	\centering
	\begin{tabular}{ccc}
		\toprule
		$g(||\nabla_x f(x))||_q)$ & WGAN & HingeGAN \\
		\cmidrule(){1-3}
		$(||\nabla_x f(x))||_1-1)^2$ & 163.2 & 179.0 \\
		$\max(0,||\nabla_x f(x))||_1-1)$ & 66.9 & 66.1 \\
		\cmidrule(){1-3}
		$(||\nabla_x f(x))||_2-1)^2$ & 34.6 & 33.8 \\
		$\max(0,||\nabla_x f(x))||_2-1)$ & 37.0 & 34.9 \\
		\cmidrule(){1-3}
		$(||\nabla_x f(x))||_{\infty}-1)^2$ & 37.5 & 33.9 \\
		$\max(0,||\nabla_x f(x))||_{\infty}-1)$ & 38.3 &  \fontseries{b}\selectfont 28.6 \\
		\cmidrule(){1-3}
		$(\text{smax}|\nabla_x f(x)|-1)^2$ & 31.9 & 283.1 \\
		$\max(0,\text{smax}|\nabla_x f(x)|-1)$ & 31.8 & 32.1 \\
		\bottomrule
	\end{tabular}
\end{table}

\begin{table}[!ht]
	\caption{FID after 100k generator iterations on CIFAR-10 using the same setting as Table1, but we are using the standard CNN architecture.}
	\label{tab:4}
	\centering
	\begin{tabular}{ccc}
		\toprule
		$g(||\nabla_x f(x))||_q)$ & WGAN & HingeGAN \\
		\cmidrule(){1-3}
		$(||\nabla_x f(x))||_2-1)^2$ & 30.2 & 26.1 \\
		$\max(0,||\nabla_x f(x))||_2-1)$ & 31.8 & 27.3 \\
		\cmidrule(){1-3}
		$\max(0,||\nabla_x f(x))||_{\infty}-1)$ & 74.3 &  \fontseries{b}\selectfont 21.3 \\
		\bottomrule
	\end{tabular}
\end{table}

\begin{table}[!ht]
	\caption{FID after 100k generator iterations on CAT (in 32x32) using the same setting as Table 1, but we are using the standard CNN architecture. Exceptionally, this set of experiment showed convergence at around 10-40k iterations (this has not been the case in any of the other experiments). For this reason, we show also the lowest FID obtained during training (FID was measured at every 10k iterations).}
	\label{tab:5}
	\centering
	\begin{tabular}{ccc}
		\toprule
		$g(||\nabla_x f(x))||_q)$ & WGAN & HingeGAN \\
		\cmidrule(){1-3}
		At 100k iterations & & \\
		\cmidrule(){1-3}
		$(||\nabla_x f(x))||_2-1)^2$ & 27.3 & \fontseries{b}\selectfont 19.5 \\
		$\max(0,||\nabla_x f(x))||_2-1)$ & 21.4 & 23.9 \\
		\cmidrule(){1-3}
		$\max(0,||\nabla_x f(x))||_{\infty}-1)$ & 66.5 & 24.0 \\
		\cmidrule(){1-3}
		Lowest FID obtained & & \\
		\cmidrule(){1-3}
		$(||\nabla_x f(x))||_2-1)^2$ & 20.9 & 16.2 \\
		$\max(0,||\nabla_x f(x))||_2-1)$ & 19.4 & 17.0 \\
		\cmidrule(){1-3}
		$\max(0,||\nabla_x f(x))||_{\infty}-1)$ & 32.32 & \fontseries{b}\selectfont 9.5 \\
		\bottomrule
	\end{tabular}
\end{table}

\begin{table}[!ht]
	\caption{FID after 100k generator iterations on CAT (in 64x64) using the same setting as Table 1.}
	\label{tab:6}
	\centering
	\begin{tabular}{ccc}
		\toprule
		$g(||\nabla_x f(x))||_q)$ & WGAN & HingeGAN \\
		\cmidrule(){1-3}
		$(||\nabla_x f(x))||_2-1)^2$ & 48.2 & 26.7 \\
		$\max(0, ||\nabla_x f(x))||_2-1)$ & 43.7 & 29.6 \\
		\cmidrule(){1-3}
		$\max(0, ||\nabla_x f(x))||_{\infty}-1)$ & 18.3 &  \fontseries{b}\selectfont 17.5 \\
		\bottomrule
	\end{tabular}
\end{table}

\begin{table}[!ht]
	\caption{Extras from Table 1}
	\label{tab:7}
	\centering
	\begin{tabular}{ccc}
		\toprule
		$g(||\nabla_x f(x))||_q)$ & WGAN & HingeGAN \\
		\cmidrule(){1-3}
		\cmidrule(){1-3}
		$(\text{smax}|\nabla_x f(x)|-1)^2$ & 35.3 & 197.9 \\
		$\max(0,\text{smax}|\nabla_x f(x)|-1)$ & 31.4 & 29.5 \\
		\bottomrule
	\end{tabular}
\end{table}

\subsection{Margins in Relativistic GANs}
\label{sec:5.5}

Relativistic paired GANs (RpGANs) and Relativistic average GANs (RaGAN) \citep{jolicoeur2018relativistic,jolicoeur2019relativistic} are GAN variants which tend to be more stable than their non-relativistic counterparts. These methods are not yet well understood and its unclear how to incorporate gradient penalty from our framework into these approach. In this subsection, we explain how we can link both approaches to MMCs.

Relativistic paired GANs (RpGANs) are defined as:
\begin{align*}
\max\limits_{C:\mathcal{X} \to \mathbb{R}} \hspace{1pt}
&\underset{ \substack{x_1 \sim \mathbb{P}  \\ z \sim \mathbb{Z}}}{\mathbb{E}\vphantom{p}} \left[ f \left( C(x_1) - C(G(z)) \right) \right], \\
\max_{G} \hspace{1pt}&
\underset{ \substack{x_1 \sim \mathbb{P}  \\ z \sim \mathbb{Z}}}{\mathbb{E}\vphantom{p}} \left[ f \left( C(G(z)) - C(x_1) \right) \right],
\end{align*}
and Relativistic average GANs (RaGANs) are defined as:
\begin{align*}
\max\limits_{C:\mathcal{X} \to \mathbb{R}} &\mathbb{E}_{x_1 \sim \mathbb{P}}\left[ f_1\left( C(x_1)-\mathbb{E}_{z \sim \mathbb{Z}} \hspace{1pt} C(G(z)) \right)) \right] + \\ &\mathbb{E}_{z \sim \mathbb{Z}} \left[ f_2 \left( C(G(z))-\mathbb{E}_{x_1 \sim \mathbb{P}} \hspace{1pt} C(x_1) \right) \right], \\
\max_{G}  \hspace{2pt} &\mathbb{E}_{z \sim \mathbb{Z}}\left[ f_1\left( C(G(z))-\mathbb{E}_{x_1 \sim \mathbb{P}} \hspace{1pt} C(x_1) \right)) \right] + \\ &\mathbb{E}_{x_1 \sim \mathbb{P}} \left[ f_2 \left( C(x_1)-\mathbb{E}_{z \sim \mathbb{Z}} \hspace{1pt} C(G(z)) \right) \right],
\end{align*}
where $f, f_1, f_2:\mathbb{R} \to \mathbb{R}$.

Most loss functions can be represented as RaGANs or RpGANs; SGAN, LSGAN, and HingeGAN all have relativistic counterparts.

\subsubsection{Relativistic average GANs}

From the loss function of RaGAN, we can deduce its decision boundary.
Contrary to typical classifiers, we define two boundaries, depending on the label.
The two surfaces are defined as two sets of points $(x_0,y_0)$ such that:
\begin{align*}
f(x_0) &= \mathbb{E}_{x \sim \mathbb{Q}}[f(x)] \text{, when } y_0 = 1 (\text{real}) \\
f(x_0) &= \mathbb{E}_{x \sim \mathbb{P}}[f(x)] \text{, when } y_0 = -1 (\text{fake})
\end{align*}
It can be shown that the relativistic average geometric margin is approximated as:
\begin{align*}
\gamma_p^{Ra}(x,y) \approx &\frac{((y+1)/2)(f(x)-\mathbb{E}_{x \sim \mathbb{Q}}[f(x)])}{|| \nabla_x f(x) ||_q} + \\ &\frac{ ((y-1)/2)(f(x)-\mathbb{E}_{x \sim \mathbb{P}}[f(x)])}{|| \nabla_x f(x) ||_q} \\ = &\frac{\alpha_{Ra}(x,y)}{\beta(x)}.
\end{align*}

Maximizing the boundary of RaGANs can be done in the following way:
\begin{align*}
\min_f \mathbb{E}_{(x,y)\sim \mathbb{D}}\left[L(\alpha_{Ra}(x,y)) + \lambda g(||\nabla_x f(x)||_q)\right].
\end{align*}

\subsubsection{Relativistic paired GANs}

From its loss function (as described in section \ref{sec:GANs}), it is not clear what the boundary of RpGANs can be. However, through reverse engineering, it is possible to realize that the boundary is the same as the one from non-relativistic GANs, but using a different margin. We previously derived that the approximated margin (non-geometric) for any point is $\gamma_p(x) \approx \frac{|f(x)|}{|| \nabla_x f(x) ||_q}$. We define the geometric margin as the margin after replacing $|f(x)|$ by $yf(x)$ so that it depends on both $x$ and $y$. However, there is an alternative way to transform the margin in order to achieve a classifier. We call it the \em relativistic paired margin\em:
\begin{align*}
\gamma^{*}_p(x_1,x_2) &= \gamma_p(x_1) - \gamma_p(x_2) \\
& \approx \frac{f(x_1)}{|| \nabla_{x_1} f(x_1) ||_q} - \frac{f(x_2)}{|| \nabla_{x_2} f(x_2) ||_q}.
\end{align*}
where $x_1$ is a sample from $\mathbb{P}$ and $x_2$ is a sample from $\mathbb{Q}$. This alternate margin does not depend on the label $y$, but only ask that for any pair of class 1 (real) and class 2 (fake) samples, we maximize the relativistic paired margin. This margin is hard to work with, but if we enforce $|| \nabla_{x_1} f(x_1) ||_q \approx || \nabla_{x_2} f(x_2) ||_q$, for all $x_1 \sim \mathbb{P}$,$x_2 \sim \mathbb{Q}$, we have that:
\begin{align*}
\gamma^{*}_p(x_1,x_2) \approx \frac{f(x_1) - f(x_2)}{|| \nabla_{x} f(x) ||_q},
\end{align*}
where $x$ is any sample (from class 1 or 2).

Thus, we can train an MMC to maximize the relativistic paired margin in the following way:
\begin{align*}
\min_f \underset{ \substack{x_1 \sim \mathbb{P}  \\ z \sim \mathbb{Z}}}{\mathbb{E}\vphantom{p}}&\left[L(f(x_1)-f(G(z)))\right] + \\ \lambda& \mathbb{E}_{(x,y)\sim \mathbb{D}}\left[g(||\nabla_x f(x)||_q)\right],
\end{align*}
where $g$ must constrains $||\nabla_x f(x)||_q$ to a constant.

This means that minimizing $L(f(x_1)-f(x_2))$ without gradient penalty can be problematic if we have different gradient norms at samples from class 1 (real) and 2 (fake). This provides an explanation as to why RpGANs do not perform very well unless using a gradient penalty \citep{jolicoeur2018relativistic}.

\subsection{Proofs}

Note that both of the following formulations represent the margin:

\begin{align*}
\gamma(x) = &\min_{x_0} || x_0 - x || \hspace{5pt} \text{ s.t. } \hspace{5pt} f(x_0)=0 \\
= &\min_{r} || r || \hspace{5pt} \text{ s.t. } \hspace{5pt} f(x+r)=0
\end{align*}

\subsubsection{Bounded Gradient $\iff$ Lipschitz }

Assume that $f:X \to \mathbb{R}$ and $X$ is a convex set.

Let $\tilde{x}(\alpha) = \alpha x_1 + (1-\alpha) x_2$, where $\alpha \in [0,1]$ be the interpolation between any two points $x_1,x_2 \in X$. We know that $\tilde{x}(\alpha) \in X$ for any $\alpha \in [0,1]$ by convexity of $X$.
\begin{align*}
f(x_1)-f(x_2) &= f(\tilde{x}(1))-f(\tilde{x}(0)) \\
&= \int_{0}^{1} \frac{df(\tilde{x}(\alpha))}{d\alpha} d\alpha \\
&= \int_{0}^{1} \nabla f(\tilde{x}(\alpha)) \frac{\tilde{x}(\alpha)}{d\alpha} d\alpha \\
&= \int_{0}^{1} \nabla f(\tilde{x}(\alpha)) (x_1-x_2) d\alpha \\
&= (x_1-x_2) \int_{0}^{1} \nabla f(\tilde{x}(\alpha)) d\alpha
\end{align*}

1) We show $||\nabla f(x)||_p \leq K \implies \frac{|f(x)-f(y)|}{||x-y||_p} \leq K$ for all $x,y$:

Let $||\nabla f(x)||_p \leq K$ for all $x \in X$.
\begin{align*}
|f(x_1)-f(x_2)| &= || (x_1-x_2) \int_{0}^{1} \nabla f(\tilde{x}(\alpha)) d\alpha ||_p \\
&\leq || x_1-x_2 ||_p \cdot ||_p \int_{0}^{1} \nabla f(\tilde{x}(\alpha)) d\alpha || \\
&\leq || x_1-x_2 ||_p \int_{0}^{1} ||\nabla f(\tilde{x}(\alpha))||_p d\alpha \\
&\leq || x_1-x_2 ||_p \int_{0}^{1} K d\alpha \\
&\leq K || x_1-x_2 ||_p \\
\end{align*}

2) We show $\frac{|f(x)-f(y)|}{||x-y||_p} \leq K$ for all $x,y$ $\implies || \nabla f(x)||_p \leq K$ for $p\ge2$:

Assume $p\ge 2$ and $\frac{1}{p}+\frac{1}{q}=1$.
\begin{align*}
|| \nabla_x f(x) ||_p &\leq || \nabla_x f(x) ||_q \hspace{5pt} \text{ since $p \ge q$} \\
&= \max_v \nabla_x f(x)^T v \hspace{5pt} \text{ s.t. } \hspace{5pt} ||v||_p \leq 1 \\
&= \left| \nabla_x f(x)^T v^{*} \right| \hspace{3pt} \text{ where $v^{*}$ is the optimum} \\
&= lim_{h\to 0} \left|  \frac{f(x+hv^{*})-f(x)}{h} \right|  \\
&\leq lim_{h\to 0} \frac{\left|f(x+hv^{*})-f(x)\right|}{h|| v^{*}  ||_p}  \\
&= lim_{h\to 0}  \frac{\left|f(x+hv^{*})-f(x)\right|}{||x +  hv^{*} - x  ||_p}  \\
&\leq K 
\end{align*}

\subsubsection{Taylor approximation}

Let $r = x_0 - x$; at the boundary $x_0$, we have that $f(x_0)=f(x+r)=C$, for some constant $C$. In the paper, we use generally assume $C=0$. We will make use of the following Taylor approximations:

\begin{align*}
f(x+r) \approx f(x) + \nabla_x f(x)^T r \\
\implies f(x) - C \approx -\nabla_x f(x)^T r
\end{align*}

and 

\begin{align*}
f(x) \approx f(x_0) + \nabla_{x_0} f(x_0)^T (x-x_0) \\
\implies f(x) - C \approx -\nabla_{x_0} f(x_0)^T r
\end{align*}

\subsubsection{Taylor approximation (After solving)}

To make things easier, we maximize $(\gamma_2(x))^2$ instead of $\gamma_2(x)$. We also maximize with respect to $x_0$ instead of $r$:

\begin{align*}
\gamma^2_2(x) &= \min_{x_0} || x_0-x ||^2_2 \hspace{5pt} \text{ s.t. } \hspace{5pt} f(x_0)=0 \\
&= \min_{x_0} || x_0-x ||^2_2 - \lambda f(x_0),
\end{align*}
where $\lambda$ is a scalar (Lagrange multiplier).
We can then differentiate with respect to $x_0$:
\begin{align}\label{eqn:8}
\nabla_{x_0} \gamma^2_2(x)  &= (x_0-x) + \lambda \nabla_{x_0} f(x_0) = 0.
\end{align}
We will then use a inner product to be able to extract the optimal Lagrange multiplier:
\begin{align*}
\implies& -\nabla_{x_0} f(x_0)^T (x_0-x) = \lambda^{*} \nabla_{x_0} f(x_0)^T \nabla_{x_0} f(x_0) \\
\implies& \lambda^{*} = -\frac{\nabla_{x_0} f(x_0)^T (x_0-x)}{\nabla_{x_0} f(x_0)^T \nabla_{x_0} f(x_0)} \\
\implies& \lambda^{*} = -\frac{\nabla_{x_0} f(x_0)^T (x_0-x) }{|| \nabla_{x_0} f(x_0) ||^2_2} 
\end{align*}
Now, we plug-in the optimal Langrange multiplier into equation \eqref{eqn:8} and we use a inner product:
\begin{align*}
\implies& x_0-x = -\lambda \nabla_{x_0} f(x_0) \\
\implies& x_0-x = \frac{\nabla_{x_0} f(x_0)^T (x_0-x) }{|| \nabla_{x_0} f(x_0) ||^2_2} \nabla_{x_0} f(x_0) \\
\implies& (x_0-x)^T (x_0-x) = \frac{\left( \nabla_{x_0} f(x_0)^T (x_0-x) \right)^2 }{|| \nabla_{x_0} f(x_0) ||^2_2} \\
\implies& \gamma^2_2(x) = || x_0 - x ||^2_2 = \frac{\left( \nabla_{x_0} f(x_0)^T (x_0-x) \right)^2 }{|| \nabla_{x_0} f(x_0) ||^2_2} \\
\implies&  \gamma_2(x) = \frac{\left| \nabla_{x_0} f(x_0)^T (x_0-x) \right|}{|| \nabla_{x_0} f(x_0) ||_2} \\
\end{align*}

\subsubsection{Taylor approximation (Before solving)}

\begin{align*}
\min_r || r ||_p& \hspace{5pt} \text{ s.t. } \hspace{5pt} f(x+r)=C \\
\approx \min_r || r ||_p& \hspace{5pt} \text{ s.t. } \hspace{5pt} f(x)+\nabla_x f(x)^T r=C \\
\implies \min_r || r ||_p& = \frac{|f(x)-C|}{\max_r \frac{|\nabla_x f(x)^T r|}{|| r ||_p}} \\ 
& = \frac{|f(x)-C|}{\max_r \frac{\nabla_x f(x)^T r}{|| r ||_p}} \\
& = \frac{|f(x)-C|}{\max_r \nabla_x f(x)^T \frac{r}{|| r ||_p}} \\ 
& = \frac{|f(x)-C|}{\max_{|| r ||_p \leq 1} \nabla_x f(x)^T r} \\ 
& = \frac{|f(x)-C|}{|| \nabla_x f(x) ||_q},
\end{align*}
where $\frac{1}{p} + \frac{1}{q} = 1$
This is true because of the definition of the Dual norm \citep{rudin1991functional}:
\[|| a ||_{*} = \max_{|| r ||_p \leq 1} a^T r = \max_r \frac{a^T r}{|| r ||_p} = ||a||_q\]

For a standard classifier, we have $C = 0$. For a RaGAN, we have $C= \mathbb{E}_{\mathbb{Q}}[f(x)]$ when $y=1$ (real) and $C= \mathbb{E}_{\mathbb{P}}[f(x)]$ when $y=-1$ (fake).

\subsection{Architectures}

\subsubsection{DCGAN 32x32 (as in Table 1, 2, 3)}

\begin{center}
\begin{tabular}{c}
	Generator \\
	\toprule\midrule
	$z \in \mathbb{R}^{128} \sim N(0,I)$ \\
	\midrule
	ConvTranspose2d 4x4, stride 1, pad 0, 128$\to$512 \\
	\midrule
	BN and ReLU \\
	\midrule
	ConvTranspose2d 4x4, stride 2, pad 1, 512$\to$256 \\
	\midrule
	BN and ReLU \\
	\midrule
	ConvTranspose2d 4x4, stride 2, pad 1, 256$\to$128 \\
	\midrule
	BN and ReLU \\
	\midrule
	ConvTranspose2d 4x4, stride 2, pad 1, 128$\to$3 \\
	\midrule
	Tanh \\
	\bottomrule
\end{tabular} 
\end{center}
\begin{center}
\begin{tabular}{c}
	Discriminator \\
	\toprule\midrule
	$x \in \mathbb{R}^{\text{3x32x32}}$ \\
	\midrule
	Conv2d 4x4, stride 2, pad 1, 3$\to$128 \\
	\midrule
	LeakyReLU 0.2 \\
	\midrule
	Conv2d 4x4, stride 2, pad 1, 128$\to$256 \\
	\midrule
	BN and LeakyReLU 0.2 \\
	\midrule
	Conv2d 4x4, stride 2, pad 1, 256$\to$512 \\
	\midrule
	BN and LeakyReLU 0.2 \\
	\midrule
	Conv2d 4x4, stride 2, pad 1, 512$\to$1 \\
	\bottomrule
\end{tabular}
\end{center}

\subsubsection{DCGAN 64x64 (as in Table 6)}

\begin{center}
\begin{tabular}{c}
	Generator \\
	\toprule\midrule
	$z \in \mathbb{R}^{128} \sim N(0,I)$ \\
	\midrule
	ConvTranspose2d 4x4, stride 1, pad 0, 128$\to$512 \\
	\midrule
	BN and ReLU \\
	\midrule
	ConvTranspose2d 4x4, stride 2, pad 1, 512$\to$256 \\
	\midrule
	BN and ReLU \\
	\midrule
	ConvTranspose2d 4x4, stride 2, pad 1, 256$\to$128 \\
	\midrule
	BN and ReLU \\
	\midrule
	ConvTranspose2d 4x4, stride 2, pad 1, 128$\to$64 \\
	\midrule
	BN and ReLU \\
	\midrule
	ConvTranspose2d 4x4, stride 2, pad 1, 64$\to$3 \\
	\midrule
	Tanh \\
	\bottomrule
\end{tabular} 
\end{center}
\begin{center}
\begin{tabular}{c}
	Discriminator \\
	\toprule\midrule
	$x \in \mathbb{R}^{\text{3x64x64}}$ \\
	\midrule
	Conv2d 4x4, stride 2, pad 1, 3$\to$64 \\
	\midrule
	LeakyReLU 0.2 \\
	\midrule
	Conv2d 4x4, stride 2, pad 1, 64$\to$128 \\
	\midrule
	BN and LeakyReLU 0.2 \\
	\midrule
	Conv2d 4x4, stride 2, pad 1, 128$\to$256 \\
	\midrule
	BN and LeakyReLU 0.2 \\
	\midrule
	Conv2d 4x4, stride 2, pad 1, 256$\to$512 \\
	\midrule
	BN and LeakyReLU 0.2 \\
	\midrule
	Conv2d 4x4, stride 2, pad 1, 512$\to$1 \\
	\bottomrule
\end{tabular}
\end{center}

\subsubsection{Standard CNN (as in Table 4, 5)}

\begin{center}
\begin{tabular}{c}
	Generator \\
	\toprule\midrule
	$z \in \mathbb{R}^{128} \sim N(0,I)$ \\
	\midrule
	linear, 128 $\to$ 512*4*4 \\
	\midrule
	Reshape, 512*4*4 $\to$ 512 x 4 x 4 \\
	\midrule
	ConvTranspose2d 4x4, stride 2, pad 1, 512$\to$256 \\
	\midrule
	BN and ReLU \\
	\midrule
	ConvTranspose2d 4x4, stride 2, pad 1, 256$\to$128 \\
	\midrule
	BN and ReLU \\
	\midrule
	ConvTranspose2d 4x4, stride 2, pad 1, 128$\to$64 \\
	\midrule
	BN and ReLU \\
	\midrule
	ConvTranspose2d 3x3, stride 1, pad 1, 64$\to$3 \\
	\midrule
	Tanh \\
	\bottomrule
\end{tabular} 
\end{center}

\begin{center}
\begin{tabular}{c}
	Discriminator \\
	\toprule\midrule
	$x \in \mathbb{R}^{\text{3x32x32}}$ \\
	\midrule
	Conv2d 3x3, stride 1, pad 1, 3$\to$64 \\
	\midrule
	LeakyReLU 0.1 \\
	\midrule
	Conv2d 4x4, stride 2, pad 1, 64$\to$64 \\
	\midrule
	LeakyReLU 0.1 \\
	\midrule
	Conv2d 3x3, stride 1, pad 1, 64$\to$128 \\
	\midrule
	LeakyReLU 0.1 \\
	\midrule
	Conv2d 4x4, stride 2, pad 1, 128$\to$128 \\
	\midrule
	LeakyReLU 0.1 \\
	\midrule
	Conv2d 3x3, stride 1, pad 1, 128$\to$256 \\
	\midrule
	LeakyReLU 0.1 \\
	\midrule
	Conv2d 4x4, stride 2, pad 1, 256$\to$256 \\
	\midrule
	LeakyReLU 0.1 \\
	\midrule
	Conv2d 3x3, stride 1, pad 1, 256$\to$512 \\
	\midrule
	Reshape, 512 x 4 x 4 $\to$ 512*4*4 \\
	\midrule
	linear, 512*4*4 $\to$ 1 \\
	\bottomrule
\end{tabular}
\end{center}

\end{document}

%% file: math_commands.tex
\usepackage{amsmath,amsfonts,bm}

\def\eqref#1{equation~\ref{#1}}

\def\1{\bm{1}}

\DeclareMathAlphabet{\mathsfit}{\encodingdefault}{\sfdefault}{m}{sl}
\SetMathAlphabet{\mathsfit}{bold}{\encodingdefault}{\sfdefault}{bx}{n}

